\documentclass[letterpaper, 10 pt, conference]{ieeeconf}  % Comment this line out if you need a4paper

\usepackage{graphicx}
\usepackage{tabularx}
\usepackage{subcaption}
\usepackage{listings}
\usepackage{mathtools}
\usepackage{multirow}
\usepackage{flushend}
\usepackage{url}
\usepackage{caption}
\IEEEoverridecommandlockouts                              % This command is only needed if 
\usepackage{epsfig} % for postscript graphics files

\title{\LARGE \bf
Collection and Evaluation of a Long-Term 4D Agri-Robotic Dataset
}

\author{Riccardo Polvara$^{*}$, Sergi Molina Mellado, Ibrahim Hroob, Grzegorz Cielniak and Marc Hanheide% <-this % stops a space
\thanks{*Corresponding author: \texttt{rpolvara@lincoln.ac.uk}}
\thanks{All the authors are within the Lincoln Centre for Autonomous Systems (LCAS), University of Lincoln, Lincoln, UK, LN6 7TS.
This work has been supported by the European Commission as part of H2020 under grant number 871704 (BACCHUS).
        }%
}

\begin{document}

\maketitle
\thispagestyle{empty}
\pagestyle{empty}

%%%%%%%%%%%%%%%%%%%%%%%%%%%%%%%%%%%%%%%%%%%%%%%%%%%%%%%%%%%%%%%%%%%%%%%%%%%%%%%%
\begin{abstract}
Long-term autonomy is one of the most demanded capabilities looked into a robot. The possibility to perform the same task over and over on a long temporal horizon, offering a high standard of reproducibility and robustness, is appealing. Long-term autonomy can play a crucial role in the adoption of robotics systems for precision agriculture, for example in assisting humans in monitoring and harvesting crops in a large orchard. With this scope in mind, we report an ongoing effort in the long-term deployment of an autonomous mobile robot in a vineyard for data collection across multiple months. The main aim is to collect data from the same area at different points in time so to be able to analyse the impact of the environmental changes in the mapping and localisation tasks.
In this work, we present a map-based localisation study taking 4 data sessions. We identify expected failures when the pre-built map visually differs from the environment's current appearance and we anticipate LTS-Net, a solution pointed at extracting stable temporal features for improving long-term 4D localisation results.
\end{abstract}

%%%%%%%%%%%%%%%%%%%%%%%%%%%%%%%%%%%%%%%%%%%%%%%%%%%%%%%%%%%%%%%%%%%%%%%%%%%%%%%%
\section{Introduction}\label{sec:introduction}
Agricultural environments, such as an orchard, offer many challenges for autonomously and safely deploying robots at work~\cite{duckett2018agricultural}. For example, the authors in \cite{10.1007/978-3-030-89177-0_17} compared state-of-the-art SLAM methods in a simulated vineyard environment and found that plants changing appearance can lead to a degradation of the localisation module, with the robot drifting from its course. The main reason why existing SLAM methods fail in guaranteeing robust and long-term performances in the agriculture domain is that they are developed and evaluated mainly in urban or indoor environments, which are consistent across time~\cite{labbe2019rtab}.  Agricultural environments, on the other hand, are subject to seasonal changes, with plants varying in size and colours, offering less stable localisation features than buildings and road infrastructures. For this reason, achieving long-term autonomy for robots in such environments is both a challenging and interesting research problem~\cite{8421618}. Motivated by this challenge, we deploy an autonomous mobile robot in the Ktima Gerovassiliou vineyard (Greece) at specific time intervals and record sensor data to better capture the map's variability and, possibly, exploit it later.
To do so, we propose a long-term robust deployment of the robot in a vineyard navigating on a topological map. We planned to record sensory data from March 2022 until the harvesting season in October 2022. Differently from ~\cite{pire2019rosario} and any other dataset present in the literature~\cite{aguiar2020localization}, our objective is not only to create a dataset of robot sensor data in the agricultural environment, but we aim at capturing and studying the long-term nature of such data.
%our objective is to create a long-term dataset of robot sensor data spanning various months which will enable a bench-marking framework for SLAM algorithms in agricultural environments.

In the remaining of this manuscript, we provide an overview of the recorded data and the scenario, how the data is stored and how it can be retrieved, an initial assessment of the degradation of the localisation algorithm over time, and we briefly discuss ongoing research toward extracting long-term temporally stable features with the scope of overcoming the shortcomings of existing LIDAR-based SLAM approaches in a continuously evolving environment.

% \section{Related Work}\label{sec:related-work}
\section{Data Collection}\label{sec:data-collection}

\subsection{Robot Setup}\label{subsec:robot-setup}
\begin{figure}
\centering
\begin{subfigure}{0.38\linewidth}
\centering
\includegraphics[width = \linewidth]{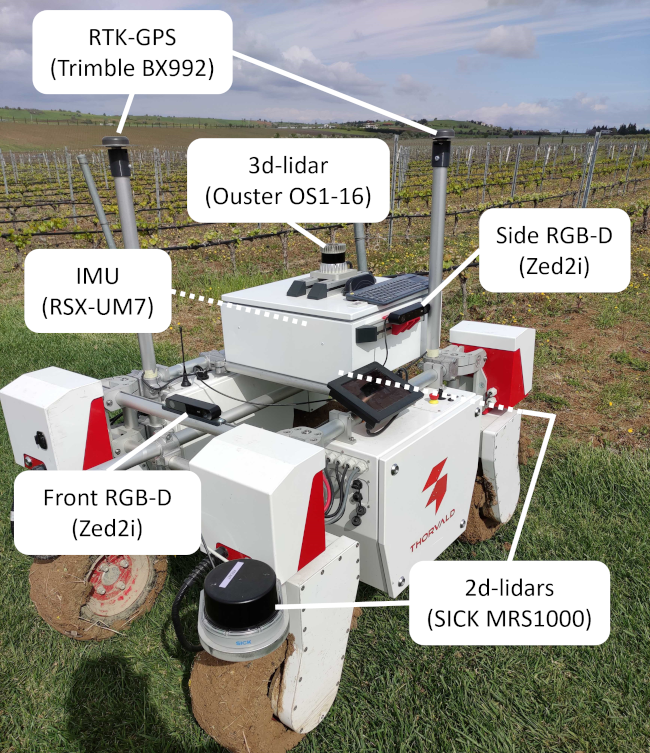}
% \caption{The Thorvald robot deployed at the Ktima Gerovassileou vineyard.}
% \label{fig:thorvald}
\end{subfigure}
\begin{subfigure}{0.58\linewidth}
\centering
\includegraphics[width = \linewidth]{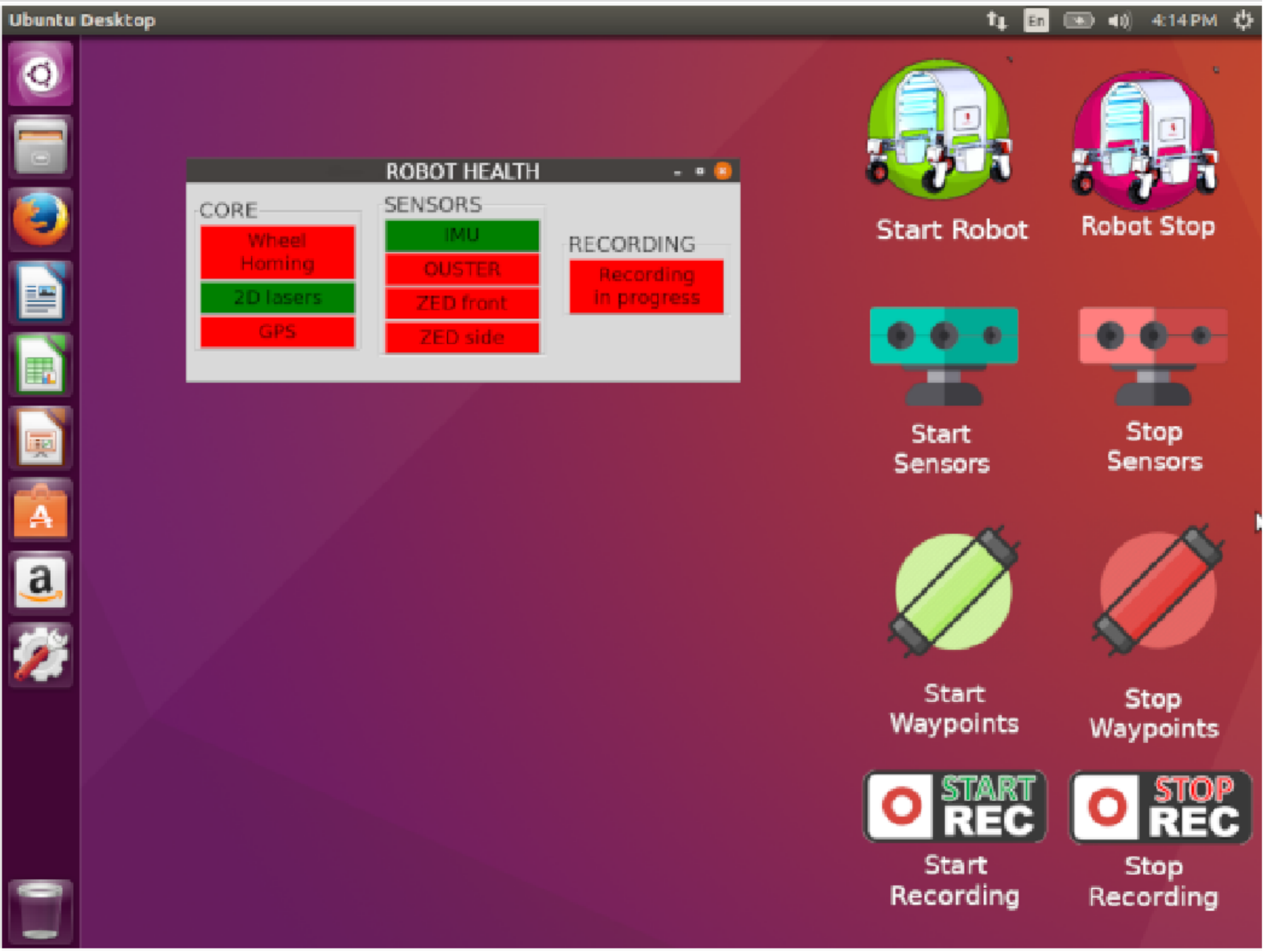}
% \caption{Special icons and GUI have been designed to facilitate the interaction with the inspection robot platform and inform the user about the current status of the modules.}
% \label{fig:gui}
\end{subfigure}
\caption{(Left) The Thorvald robot deployed at the Ktima Gerovassiliou vineyard. (Right) Special icons and GUI have been designed to facilitate the interaction with the robot.}
\label{fig:combined}
\end{figure}

For the scope of data collection, we are using the Thorvald robot (SAGA Robotics). It is equipped with a complete sensor suite consisting of an RTK-GPS (Trimble BX992), two RGBD cameras (Stereolabs ZED2i), one 16-beams 3D LIDAR (OUSTER OS1-16), IMU (RSX-UM7) and two 4-beams 2D LIDARs (SICK  MRS1000) located at opposite corners of the robot. Figure~\ref{fig:combined}(left) offers a visual representation of the robot and its sensor suite.

In order to simplify the operation of the robot and make it more accessible also to inexperienced users, special desktop icons and a GUI have been designed (Figure~\ref{fig:combined}(right)), while a touch monitor is mounted on the robot.
More specifically, two icons ``Start Robot'' and ``Robot Stop'' allows to launch and stop, respectively, the basic functionalities of the platform, such as the \textit{roscore}, the 2D lasers for obstacle detection, the RTK-GPS for outdoor localisation, and the topological\_navigation\cite{toponav} module for navigating the rows. When starting the robot, the \textit{ROBOT HEALTH} module is also launched to inform the user about the platform's status. A red/green palette is used to communicate if a specific module is currently working or not. %For example, in Figure~\ref{fig:combined}(right) is possible to see how the 2D Lasers light is green meaning the lasers are correctly up and running. However, the GPS label is still red indicating the GPS is not operating or the signal error is above a threshold, hence it would not be safe to operate the robot in the current situation.
Specials icons allow to start/stop the sensor suite (IMU, 3D LIDAR, and 2 RGBD cameras) plugged into the additional computer. In an analogous way, icons have been designed for starting the data recording session and the navigation task. %An instruction video has been realised to support any new user in approaching the inspection platform for the first time. The video is hosted on the UL's YouTube channel.

% \begin{figure}
%   \includegraphics[width=\linewidth]{pictures/thorvald_gui.pdf}
%   \caption{Special icons and GUI have been designed to facilitate the interaction with the inspection robot platform and inform the user about the current status of the modules.}
%   \label{fig:gui}
% \end{figure}

\begin{figure}
  \includegraphics[width=\linewidth]{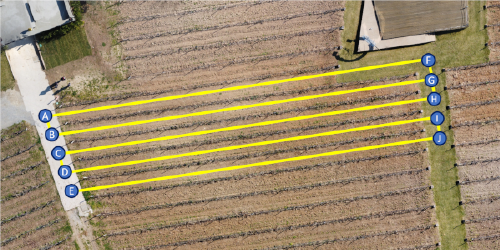}
  \caption{Topological map with the nodes and the edges followed by the robot during the data collection.}
  \label{fig:path}
\end{figure}

The area used for data collection includes five adjacent rows within the vineyard Ktima Gerovassiliou, Greece (lat: 40.450135, lon: 22.924247). By pressing the ``Start Waypoints'' icon, the robot traverses autonomously a human-designed path along the edges connecting the nodes of a topological map (Figure~\ref{fig:path}). The path defined is kept the same for all data collection sessions and is the following: $B \rightarrow G \rightarrow F \rightarrow A \rightarrow B \rightarrow G \rightarrow H \rightarrow C \rightarrow D \rightarrow I \rightarrow J \rightarrow E \rightarrow J \rightarrow I \rightarrow D \rightarrow C \rightarrow H \rightarrow G \rightarrow B$. The length of this entire path is approximately $500$\,m, and it takes, on average, $25$\,min for the robot to complete the path. The average robot speed during the traversal of a row is $0.6$\,m/s.

\subsection{Data Recording and Storage}\label{subsec:data-recording}

\begin{figure*}
\centering
\begin{tabular}{cccc}
\includegraphics[width=0.24\textwidth]{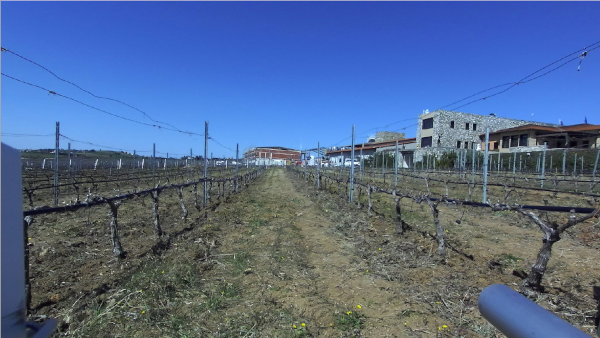} \includegraphics[width=0.24\textwidth]{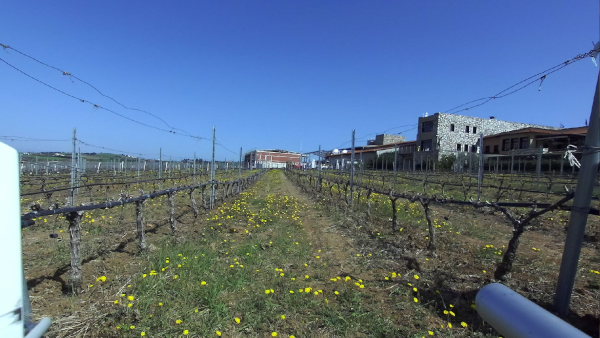}
\includegraphics[width=0.24\textwidth]{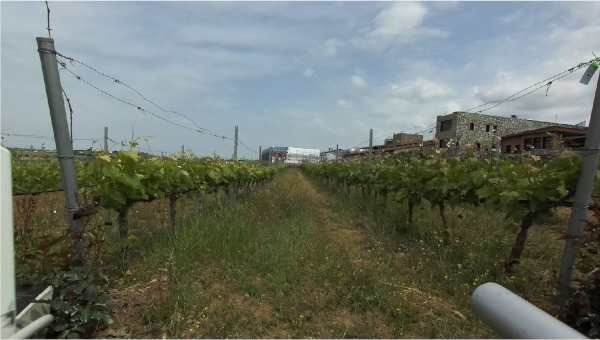} \includegraphics[width=0.24\textwidth]{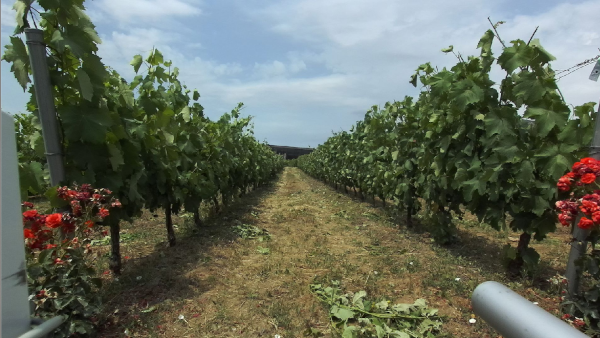}
% (a) first & (b) second \\
 
% (c) third & (d) fourth 
\end{tabular}
\caption{Different crop's growth stage starting from March (left) to June (right).}
\label{fig:crop-stages}
\end{figure*}

Our objective is to collect data until the end of the harvesting season to capture the entire crops’ growth. This process started in March, and it is currently in progress. The data recording is happening at a different frequency in order to accommodate a variable speed in the crop’s growth stages, e.g., only one session has been performed in March, while four sessions (one week apart one from another) have been completed in June. This is because during winter the plants grow very slowly, not offering significant visual changes between consecutive weeks. On the other hand, in the warmest months of the year, the plants drastically change their aspect a few days apart. A comparison between how the plants look from March to June (left to right) is offered in Figure~\ref{fig:crop-stages}.

Data is recorded and stored in a MongoDB database by using topic\_store (TS)\cite{Kirk_topic_store_2020}. Unlike ROS bags, TS adds flexibility by serialising all the ROS messages into a data hierarchy that is easily searchable with database queries and allows for remote storage. The user simply needs to define a scenario file listing all the topics to record and the hierarchy to adopt. 
% The scenario file used for recording data at Ktima Gerovassiliou is shown in Figure 8. Here, it is possible to see that 
In our setup, all the sensory information (2D/3D LIDARs, RGBD, RTK-GPS and IMU) is going to be saved, together with other navigation data such as motor signals, odometry, and current location in the topological map. 
 
% Figure 8. Scenario configuration file for recording data with topic\_store at Ktima Gerovassiliou.
Data is stored and indexed locally and then synchronised with a remote server. This can be queried and inspected to retrieve data at a specific time and location.
%MongoDB is an open source, non relational database management system (DBMS) that uses flexible documents instead of tables and rows to process and store various forms of data. As a NoSQL solution, MongoDB does not require a relational database management system (RDBMS), so it provides an elastic data storage model that enables users to store and query multivariate data types with ease.
%MongoDB documents or collections of documents are the basic units of data. Formatted as Binary JSON (Java Script Object Notation), these documents can store several types of data and be distributed across multiple systems.
The most-user-friendly way to inspect the database is by using MongoDB \textit{Compass} and establishing a working connection to the database itself. %For inspecting the local database acquired by the inspection platform at the end of the recording session, the user only needs to specify the correct hostname and port. % as in Figure 9.
 
% Figure 9. Connection details for a local MongoDB instance.

\begin{figure}
  \includegraphics[width=\linewidth]{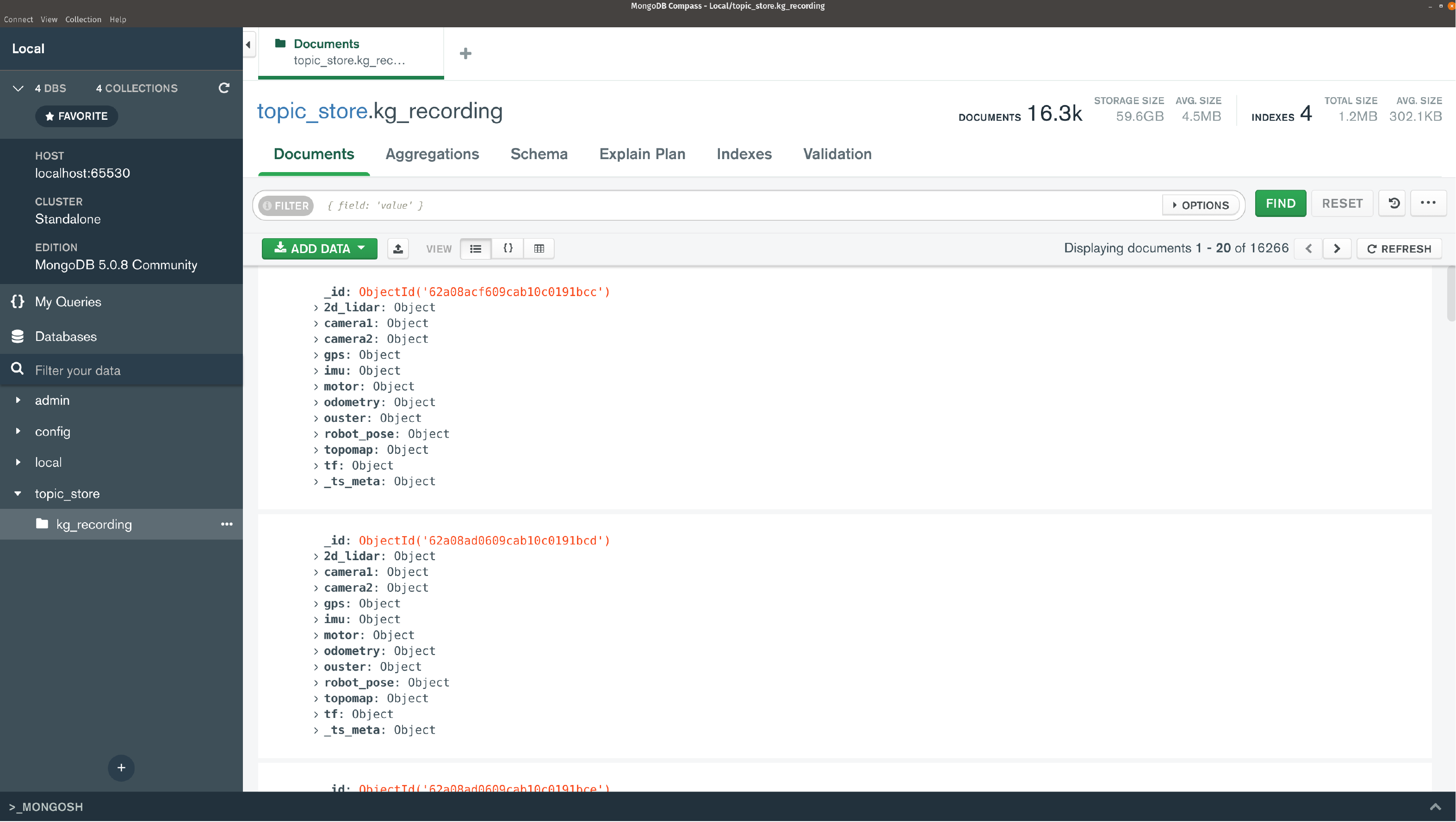}
  \caption{Overview of MongoDB Compass for inspecting the data collected during the recording session.}
  \label{fig:compass}
\end{figure}

Once the connection to the server has been established, it is possible to have access to all the databases and their respective collections. By selecting “kg\_recording” we can have an overview of the documents that have been created during the data collection session, as seen in Figure~\ref{fig:compass}. %A general overview of how MongoDB Compass looks like is offered in Figure~\ref{fig:compass}.
Each document has the same hierarchical structure defined in the TS scenario file. %, as can be seen in Figure 11. By clicking on a single document in MongoDB Compass, we can expand this structure and visualize the value of each respective field. An example is offered in Figure 11 and Figure 12.

\subsection{Database connection}\label{subsec:db-connection}
To connect to the remote server hosting the data collected across all the recording sessions, a generic user needs first to define a few environmental variables such as the database's and collection's name but also the user credentials.
A dedicated script allows extracting data from the remote database in the format of a ROS bag. In order to do so, a user needs, first of all, to define a particular query so as to filter only a portion of data. For example, a user would be interested to extract only a specific recording session out of the complete set of data. Hence, after having identified the \textit{ObjectId} of the session they are interested in (e.g, 03032022), they can type in the command line the following:

\begin{lstlisting}[language=bash, breaklines=true]
rosrun topic_store convert.py -i "mongodb://$MONGOUSER:$MONGOPASS@files.lar.lincoln.ac.uk:8098/?authSource=bacchus&tls=true&tlsInsecure=true" \
    -c kg_recording  \
    -q '{"_ts_meta.session":"ObjectId(03032022)"}' \
    -o session_03032022.bag
\end{lstlisting}
One can also add a projection (-p) to select which fields should be extracted (e.g., only the laser data).

We identified the lack of a standard benchmark for mobile robots deployed in agriculture domains as one of the biggest challenges faced by the robotic community. Hence, it is our plan to publicly release the dataset under an open-source license as soon as we have enough significant data.
Even though the primary application for this dataset is to evaluate localisation algorithms under various changing conditions, we believe that thanks to its temporal aspect, the dataset can also be used for phenotyping and crop mapping tasks (e.g. see \cite{lu2020survey}).

\section{Data evaluation}\label{sec:data-evaluation}

\begin{figure}
  \includegraphics[width=\linewidth]{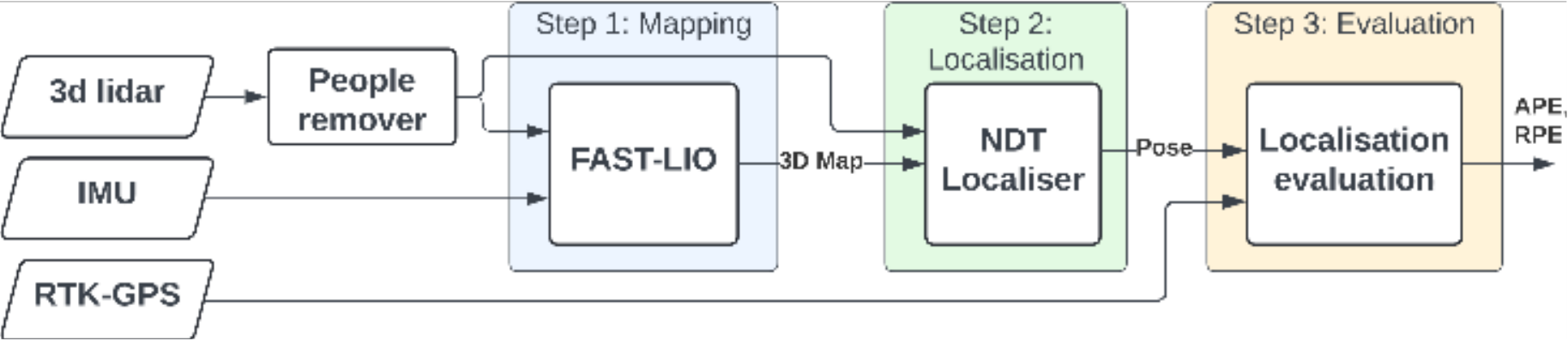}
  \caption{Pipeline followed for the localisation evaluation.}
  \label{fig:localization-pipeline}
\end{figure}

As just said, the final objective of the data collection is to create a dataset of robot sensor data that can be useful for developing and bench-marking novel navigation and localisation algorithms in the agricultural domain. An interesting research question to address is about the long-term deployment of a robot in an environment that is subject to seasonal changes, offering severe challenges for what concerns the robot’s localisation in such an environment. 
Within this scope, we are providing in this manuscript a comparison performed using data from four different data recording sessions, the same from which we extracted the frames shown in Figure~\ref{fig:crop-stages}. To be specific, the recording sessions occurred on the 3rd March 2022, 6th April 2022, 6th May 2022, and 8th June 2022.
The goal of this study is to quantify how and if the localisation performance drops over time when trying to localise the robot by using a map built in a different moment, e.g., the robot aims at localising itself in April while using a map built in March.
To do so, we follow a 3-step process as shown in Figure~\ref{fig:localization-pipeline}, consisting of mapping, localisation, and evaluation. 
Before the 3D LIDAR point cloud is fed in the FAST-LIO~\cite{xu2021fast} mapping and NDT-Localizer~\cite{magnusson2009three} algorithms, a people remover filter step is applied to the raw points. For safety reasons, during the recording process, a person is always following the robot some meters behind in case an emergency stop must be triggered. So, in order to remove the person, a 3D area behind the robot is cropped out of all the scans using the pcl library.

\subsection{3D Mapping}\label{subsec:mapping}
To this scope, we initially create a 3D map using FAST-LIO (Fast LiDAR-Inertial Odometry). This uses as input the point cloud data from the Ouster OS1 3D LIDAR mounted on top of our robot and its IMU data. FAST-LIO is a computationally efficient and robust LIDAR-inertial odometry package. It fuses LIDAR feature points with IMU data using a tightly coupled iterated extended Kalman filter to allow robust navigation in fast-motion, noisy or cluttered environments where degeneration occurs. The resulting output is a 3D point cloud in pcd format. 

% \begin{figure}
% \centering
% \begin{subfigure}{0.49\linewidth}
% \centering
% \includegraphics[width =\linewidth]{pictures/pcl/march_map.png}
% % \caption{The Thorvald robot deployed at the Ktima Gerovassileou vineyard.}
% \label{fig:map-march}
% \end{subfigure}
% \begin{subfigure}{0.49\linewidth}
% \centering
% \includegraphics[width = \linewidth]{pictures/pcl/june_map.png}
% % \caption{Special icons and GUI have been designed to facilitate the interaction with the inspection robot platform and inform the user about the current status of the modules.}
% \label{fig:map-june}
% \end{subfigure}
% \caption{Point cloud map of KG built by FAST-LIO out of the data recorded in March 2022 (left) and in June 2022 (right).}
% \label{fig:map-march-june}
% \end{figure}

\begin{figure}
\centering
\begin{tabularx}{\columnwidth}{XX}
\includegraphics[width=0.46\columnwidth]{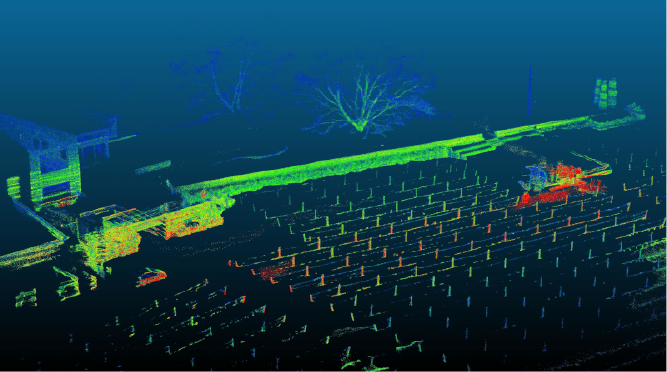} 
\includegraphics[width=0.46\columnwidth]{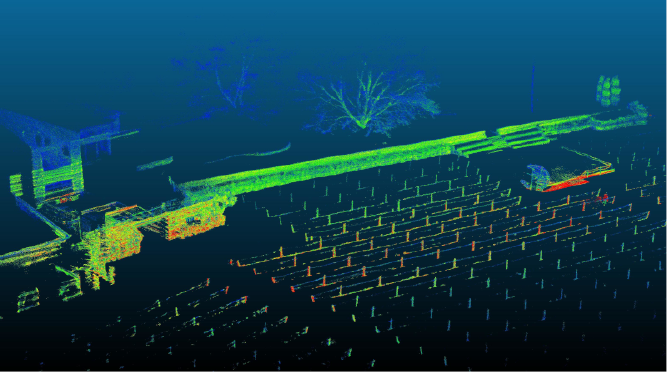} \\
\includegraphics[width=0.46\columnwidth]{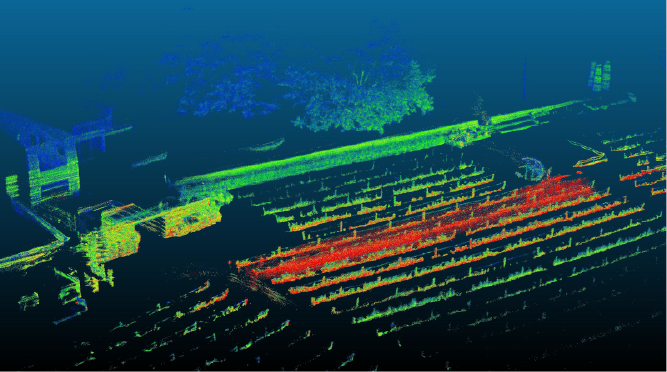} 
\includegraphics[width=0.46\columnwidth]{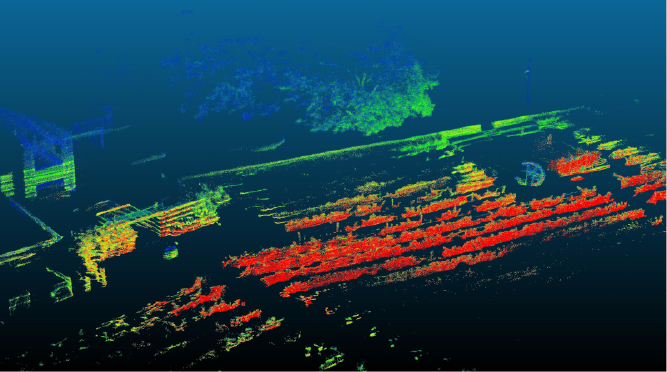}
\end{tabularx}
\caption{Row mapping detail starting from March(top-left) to June(bottom-right). Moving from late winter across spring and into summer, poles disappear from the map because the plants are now covered by leaves, which also occlude LIDAR beams preventing the mapping of the surrounding environment.}
\label{fig:pcl-maps}
\end{figure}

% Two examples of maps are shown in Figure~\ref{fig:map-march-june}, built by using FAST-LIO on data collected in March and June while traversing a single row. For the second map, it is possible to see that the lush plants located at the two sides of the robot do create occlusion to the LIDAR beams, preventing other rows and buildings to be mapped at all. 
A visual representation of the 4 maps obtained using FAST-LIO for the four sessions is offered in Figure~\ref{fig:pcl-maps}. %, with a zoomed-in detail illustrated in Figure~\ref{fig: pcl-maps-details}. 
Moving from late winter across spring and into summer, poles disappear from the map because the plants are now covered by leaves, which also occlude lidar beams preventing mapping more of the surrounding environment.

\subsection{Localisation}\label{subsec:localization}
The resulting maps are then used for localising the robot using a method based on the Normal Distribution Transform (NDT), which tries to match each point cloud obtained by the 3D LIDAR with the pre-recorded map. However, this technique does not try to match the current lidar scan to points in the map; instead, it tries to match points from the current scan to a 3D grid of probability functions created from the map. The probability distributions are obtained by dividing the point cloud into a uniform 3D grid where each 3D cell (voxel) contains the mean and standard deviation of the points belonging to it, representing a normal distribution. This way, even if the robot measures a point a few millimetres away from where the map thinks a point should be, instead of being completely unable to match those two points, the NDT matching function connects the measured point to the probability function on the map. 
There are 3 factors worth mentioning which affect the performance of the NDT localizer: (i) the down-sampling method of the input point cloud (voxel grid filter), (ii) the resolution of the point cloud map, and (iii) the initial pose given to the robot at the beginning of the process. Regarding (i), all the points present in each voxel of the input voxel grid will be approximated (i.e., down-sampled) with their centroid. This approach is a bit slower than approximating them with the centre of the voxel, but it represents the underlying surface more accurately. An experimental analysis found that a leaf (voxel side's length) of $0.1$\,m offers the best trade-off between computational efficiency and properly capturing salient features in the map, such as leaves and crops. The resolution parameter (ii) defines the voxel resolution of the internal NDT grid structure used for representing the environmental map. This structure is easily searchable, and each voxel contains the statistical data (e.g., mean and covariance) associated with the points it contains. The statistical data is used to model the cloud as a set of multivariate Gaussian distributions and allows us to calculate and optimise the probability of the existence of points at any position within the voxel. This parameter needs to be large enough for each voxel to contain at least six points but small enough to uniquely describe the environment. For this reason, we found a value of $2.0$ to work for our setting. Finally, for what concerns the initial pose of the robot (iii), in all the experiments we set it to be the same from which the mapping task has started, so as to facilitate the initial localisation of the robot itself. In fact, all the ROS bags have been extracted from the database by filtering for the same starting position (in GPS coordinates) so as to spatially align them in an easy way.

\subsection{Evaluation}\label{subsec:evaluation}
In order to evaluate the NDT-Localizer performance, we compare the output produced as estimated odometry against the pose given by the RTK-GPS. We define a grid-matrix experiment in which, for every month's map, we compute the localisation error while playing back the ROS bag of every data recording session for all the months. 
As a testbed, we assessed the robot traversing the first two rows of the research area at Ktima Gerovassiliou. More specifically, the robot starts at the beginning of the second row, which is traversed entirely, and then returns it from the first row before traversing the second row again, completing a loop. Taking as reference Figure~\ref{fig:path}, the path is $B \rightarrow G \rightarrow F \rightarrow A \rightarrow B \rightarrow G$. In total, the traversed distance is $~150$ meters. 
  
\subsubsection{Metrics}
The metrics used for evaluating the localisation errors are the Absolute Pose Error (APE), the Relative Pose Error (RPE), and the Root Mean Squared Error (RMSE). 
Regarding APE, corresponding poses are directly compared between estimate and reference, given a pose relation. Then, statistics for the whole trajectory are calculated. This is useful to test the global consistency of a trajectory. APE is based on the absolute relative pose between two poses $P_{est,i}, P_{ref,i} \in SE(3)$ at timestamp $i$ \cite{lu1997globally}:
\begin{equation}
    E_i=P_{est,i} \ominus P_{ref,i} = P_{ref,i}^{-1}  P_{est,i} \in SE(3)
\end{equation}
Where $\ominus$ is the inverse composition operator, which takes two poses and gives the relative pose. Different pose relations can be used to calculate the APE, such as only translation, only rotation, or the full transformation:
\begin{equation}
    APE_i= \parallel E_i - I_{4\times4} \parallel_F    
\end{equation}

RPE compares instead the relative poses along the estimated and the reference trajectory. This is based on the delta pose difference. This metric gives insights into the local accuracy, i.e., the drift.

\begin{equation}
\begin{split}
    E_{i,j} &= \delta_{est_{i,j}} \ominus \delta_{ref_{i,j}} \\
            &= (P_{ref,i}^{-1} P_{ref,j})(P_{est,i}^{-1},P_{est,i} ) \in SE(3)    
\end{split}
\end{equation}    

It is possible to use different pose relations to calculate the RPE from timestamp \textit{i} to \textit{j}. For example, if we want to consider the full transformation as before:
\begin{equation}
    RPE_{i,j}= \parallel E_{i,j} - I_{4\times4} \parallel_F    
\end{equation}

Then, different statistics can be calculated on the APEs and RPEs of all timestamps, e.g., the RMSE:

% \noindent\begin{minipage}{.5\linewidth}
\begin{equation}
  RMSE = \sqrt{\frac{1}{N} \sum_{\forall_{ij}} APE_{i,j}^2 }
\end{equation}
% \end{minipage}%
% or 
% \begin{minipage}{.5\linewidth}
% \begin{equation}
%   RMSE = \sqrt{\frac{1}{N} \sum_{\forall_{ij}} RPE_{i,j}^2 } 
% \end{equation}
% \end{minipage}

The Evo~\cite{grupp2017evo} library has been used for computing the metrics and comparing the resulting trajectories. 

\subsubsection{Results}

\begin{table*}[!ht]
\caption{Loop Experiment: Absolute Pose Error (APE – expressed as mean and standard deviation) and Root Mean Square Error (RMSE) while cross-testing the NDT-Localizer on seasonal data. The best result is in bold.}
\label{tab:ape}
\centering
\begin{tabular}{cc|cccccccc}
\hline
\multirow{2}{*}{}                         & \multirow{2}{*}{BAG} & \multicolumn{2}{c}{March} & \multicolumn{2}{c}{April}           & \multicolumn{2}{c}{May}             & \multicolumn{2}{c}{June}            \\
                                          &                      & APE            & RMSE     & APE                 & RMSE          & APE                 & RMSE          & APE                 & RMSE          \\ \hline
\multirow{2}{*}{MAP} & \multicolumn{1}{c|}{March}                & 0.31(0.17)     & 0.36     & \textbf{0.29(0.18)} & \textbf{0.35} & 0.35(0.20)          & 0.40          & 0.44(0.22)          & 0.49          \\
                    & \multicolumn{1}{c|}{April}                & 0.51(0.29)     & 0.59     & \textbf{0.37(0.19)} & \textbf{0.42} & 0.39(0.20)          & 0.44          & 0.50(0.52)          & 0.72          \\
                    & \multicolumn{1}{c|}{May}                  & 0.40(0.31)     & 0.51     & 0.41(0.16)          & 0.43          & \textbf{0.24(0.21)} & \textbf{0.32} & 0.23(0.43)          & 0.48          \\
                    & \multicolumn{1}{c|}{June}                 & --             & --       & --                  & --            & 0.22(0.08)          & 0.25          & \textbf{0.16(0.07)} & \textbf{0.17} \\ \hline
\end{tabular}
\end{table*}

\begin{table*}[!ht]
\caption{Loop Experiment: Relative Pose Error (RPE – expressed as mean and standard deviation) and Root Mean Square Error (RMSE) while cross-testing the NDT-Localizer on seasonal data. The best result is in bold.}
\label{tab:rpe}
\centering
\begin{tabular}{cccccccccc}
\hline
\multirow{2}{*}{}                         & \multirow{2}{*}{BAG} & \multicolumn{2}{c}{March}           & \multicolumn{2}{c}{April}           & \multicolumn{2}{c}{May} & \multicolumn{2}{c}{June}            \\
                                          &                      & RPE                 & RMSE          & RPE                 & RMSE          & RPE           & RMSE    & RPE                 & RMSE          \\ \hline
\multirow{2}{*}{MAP} & \multicolumn{1}{c|}{March}                & \textbf{0.03(0.01)} & \textbf{0.12} & 0.03(0.03)          & 0.04          & 0.03(0.06)    & 0.07    & 0.05(0.06)          & 0.08          \\
                     & \multicolumn{1}{c|}{April}                & 0.04(0.14)          & 0.14          & \textbf{0.02(0.06)} & \textbf{0.07} & 0.03(0.02)    & 0.04    & 0.06(0.16)          & 0.17          \\
                     & \multicolumn{1}{c|}{May}                  & 0.05(0.19)          & 0.19          & \textbf{0.03(0.02)} & \textbf{0.03} & 0.05(0.02)    & 0.03    & 0.05(0.33)          & 0.34          \\
                     & \multicolumn{1}{c|}{June}                 & --                  & --            & --                  & --            & 0.03(0.07)    & 0.08    & \textbf{0.03(0.04)} & \textbf{0.05} \\ \hline
\end{tabular}
\end{table*}

The results are reported Table~\ref{tab:ape} and Table~\ref{tab:rpe} for the APE and the RPE, respectively. It is possible to see how, in general, the localisation error grows in those situations in which the map and the ROS bag belong to different months. For example, the first row of Table~\ref{tab:ape} shows that, while localising on the map built in March, the localisation error for May ($0.35$) and June ($0.44$) is bigger than the one for March and April (which are comparable – $~0.30$). More specifically, it can be noted that the smallest error is always located on the main diagonal of the resulting matrix, meaning that localising a ROS bag on the map built out of the same bag always leads to better performance. In other words, localising on a map built in the same month in which the robot is deployed (or around the same date) achieves the lowest localisation error.

The most interesting result obtained, but not entirely unexpected, is that testing the ROS bags belonging to the March and April sessions on the summer map of June does fail to the point that metrics could not be computed. An intuitive explanation of such failure is that the sensor readings in March and April are quite sparse because the vines are still in a “dormant” stage and no foliage is present on them. Therefore, the only features present in the point cloud belong to the buildings and the structure which is located far from the robot. However, these features are not present in June's map due to occlusions. In fact, during the summer months, the crops are in a growth stage so advanced that almost prevents any laser beam to leave the row in which the robot is located, impacting the mapping quality of the surrounding area.

\subsection{Adding Long-Term Stability Mapping}\label{subsec:long-term}
Capturing only the stable features in a given environment is the essence of enabling long-term operation. With this objective in mind, we propose LTS-Net\footnote{This work is currently under review.}, an end-to-end self-supervised learning model. The approach is data-driven and learns implicitly the long-term stable features by means of a neural network based on PointNet++~\cite{qi2017pointnet++}. The training data is generated in an unsupervised way by having multiple observations of the same environment taken at different times, and producing a pointwise labelling representing each point's long-term stability. The network is a regression model trained by using the stability score as a training signal. This approach was initially evaluated in the NCLT dataset~\cite{carlevaris2016university}, which provides data over multiple months. The results showed that the network is able to implicitly learn the long-term stable features of that dataset. However, the nature of the NCTL dataset, collected inside a university campus and characterised by a large number of temporally stable features such as buildings and street lamps, offers limited possibilities to learn the long-term stability compared to the dataset we are proposing.

\begin{table*}[!ht]
\caption{Loop Experiment: Absolute Pose Error (APE – expressed as mean and standard deviation) and Root Mean Square Error (RMSE) while cross testing the NDT-Localizer on seasonal data and map cleaned by LTS-Net. The best result is in bold.}
\label{tab:stability-ape}
\centering
\begin{tabular}{cccccccccc}
\hline
\multirow{2}{*}{}                         & \multirow{2}{*}{BAG} & \multicolumn{2}{c}{March} & \multicolumn{2}{c}{April}           & \multicolumn{2}{c}{May}             & \multicolumn{2}{c}{June}            \\
                                          &                      & APE            & RMSE     & APE                 & RMSE          & APE                 & RMSE          & APE                 & RMSE          \\ \hline
\multirow{2}{*}{MAP} & \multicolumn{1}{c|}{June}                & --     & -- & -- & --    & 0.22(0.08) & 0.25 & 0.16(0.07)           & 0.17         \\
& \multicolumn{1}{c|}{June-filtered}                 & --     & --     & \textbf{0.16(0.06) } & \textbf{0.17 } & \textbf{0.20(0.05) }          & \textbf{0.21}         & \textbf{0.07(0.04) }          & \textbf{0.08}          \\ \hline
\end{tabular}
\end{table*}

\begin{table*}[!ht]
\caption{Loop Experiment: Relative Pose Error (RPE – expressed as mean and standard deviation) and Root Mean Square Error (RMSE) while cross testing the NDT-Localizer on seasonal data and map cleaned by LTS-Net. The best result is in bold.}
\label{tab:stability-rpe}
\centering
\begin{tabular}{cccccccccc}
\hline
\multirow{2}{*}{}                         & \multirow{2}{*}{BAG} & \multicolumn{2}{c}{March} & \multicolumn{2}{c}{April}           & \multicolumn{2}{c}{May}             & \multicolumn{2}{c}{June}            \\
                                          &                      & APE            & RMSE     & APE                 & RMSE          & APE                 & RMSE          & APE                 & RMSE          \\ \hline
\multirow{2}{*}{MAP} & \multicolumn{1}{c|}{June}                & --     & -- & -- & --    & 0.03(0.07)  & 0.08 & \textbf{0.03(0.04)}            & 0.05          \\
                    & \multicolumn{1}{c|}{June-filtered}                & --     & --     & \textbf{0.02(0.03)  } & \textbf{0.04 } & \textbf{0.02(0.02) ) }          & \textbf{0.03}         & 0.04(0.05)           & 0.06         \\ \hline
\end{tabular}
\end{table*}

\begin{figure}
  \includegraphics[width=\linewidth]{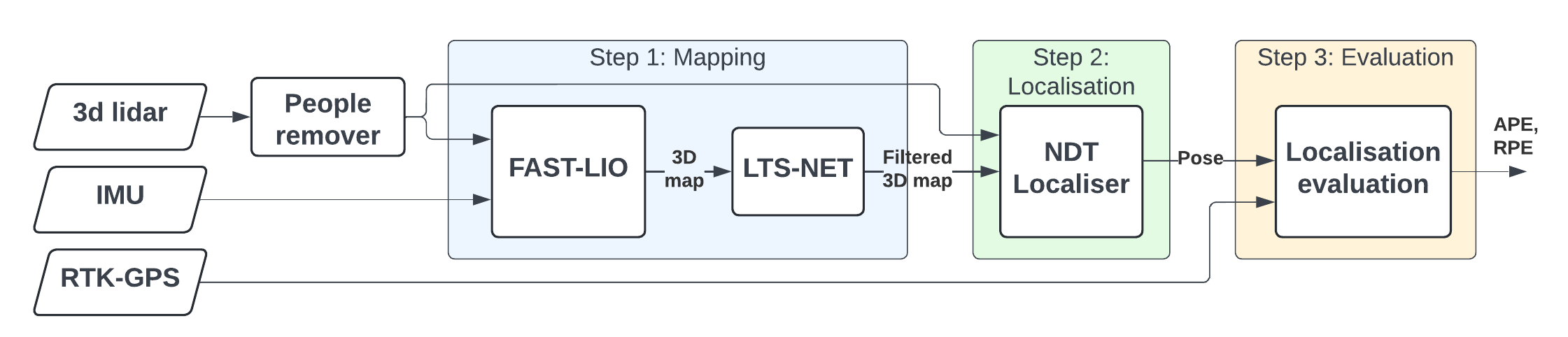}
  \caption{Evaluation pipeline with the introduction of the LTS-Net }
  \label{fig:localization-pipeline-tsnet}
\end{figure}

% \begin{figure}
% \centering
% \setkeys{Gin}{width=\columnwidth}
% \begin{tabularx}{\columnwidth}{cc}
% \includegraphics[width=0.45\columnwidth]{pictures/stability/june_unstable.pdf} 
% \includegraphics[width=0.45\columnwidth]{pictures/stability/june_stable.pdf} \\
% \includegraphics[width=0.45\columnwidth]{pictures/stability/june_detail_unstable.pdf} 
% \includegraphics[width=0.45\columnwidth]{pictures/stability/june_detail_stable.pdf}
% \end{tabularx}
% \caption{Visual comparison of stable(blue) and unstable (green to red) features in the map built out of the data recording session of June before(left) and after(right) the filtering performed by LTS-Net. }
% \label{fig:stability-comparison}
% \end{figure}

\begin{figure}
\centering
\setkeys{Gin}{width=\columnwidth}
\begin{tabularx}{\columnwidth}{cc}
\includegraphics[width=0.47\columnwidth]{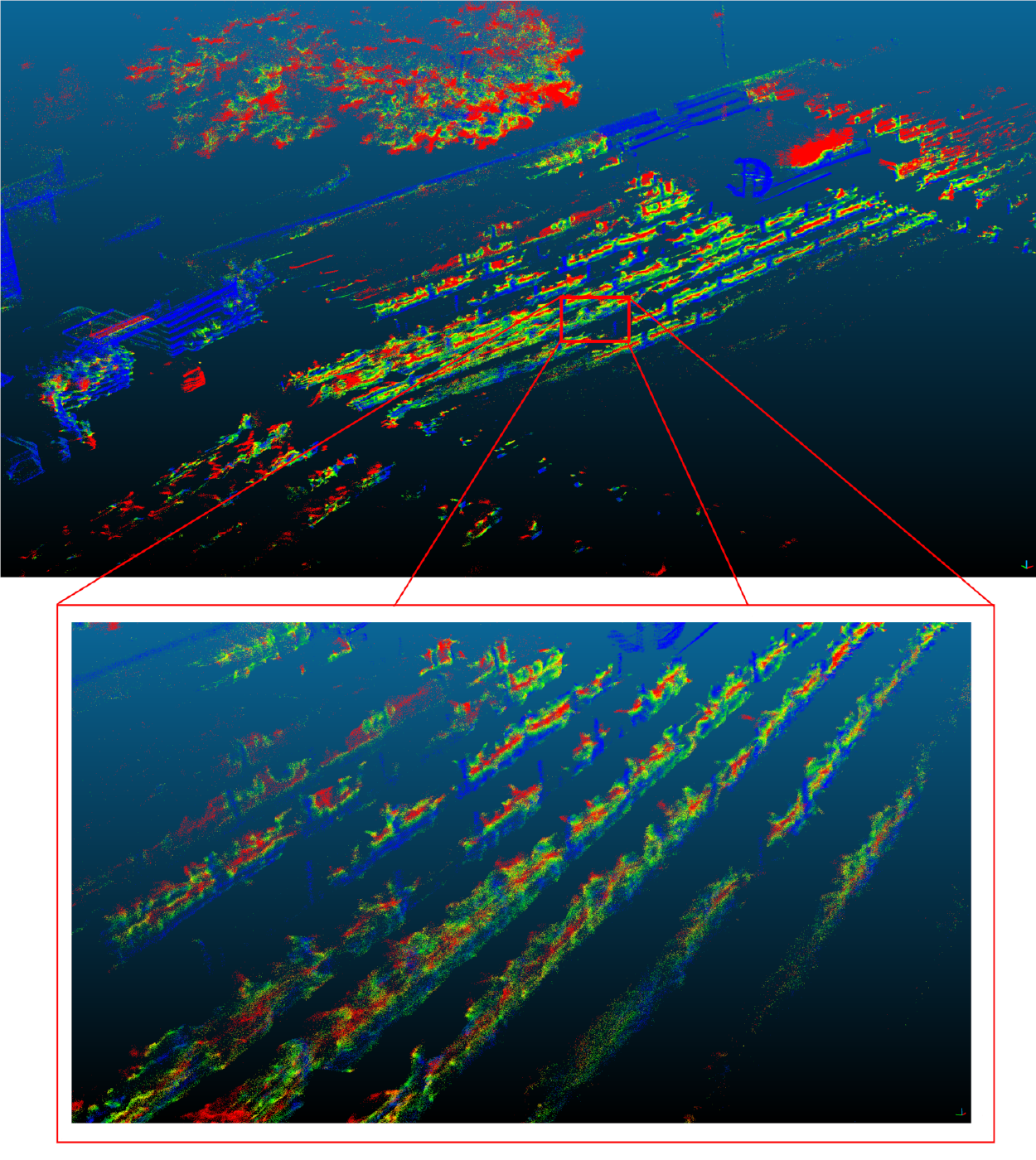} 
\includegraphics[width=0.47\columnwidth]{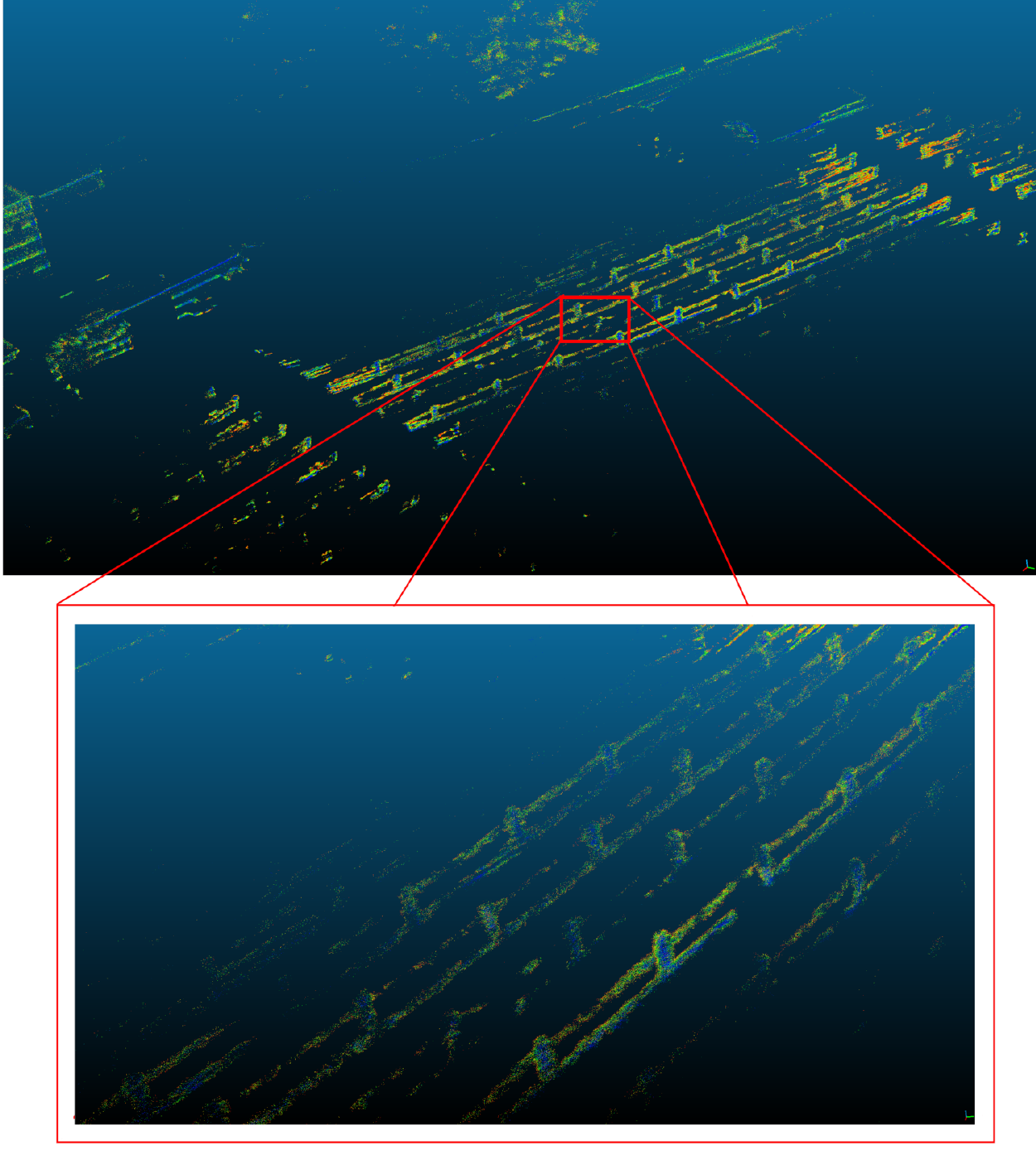}
\end{tabularx}
\caption{Visual comparison of stable(blue) and unstable (green to red) features in the map built out of the data recording session of June before(left) and after(right) the filtering performed by LTS-Net. }
\label{fig:stability-comparison}
\end{figure}

Therefore, we repeat the same experiment done in the previous section using the map from June after being filtered of the unstable points by using the target signal which LTS-Net is trained with (see the update pipeline in Figure~\ref{fig:localization-pipeline-tsnet}). A visual comparison of the original map against the ground-truth filtered one is presented in Figure~\ref{fig:stability-comparison} (the ground points have been removed for clarity). The results of the experiments are reported in Table~\ref{tab:stability-ape} and Table~\ref{tab:stability-rpe}, in which we also report the metrics obtained using the original map to facilitate the comparison. 

This final experiment shows that, by removing the temporally unstable points, is now possible to localise the ROS bag recorded in April on June's map, with a discretely low error too ($0.16$). This validates the idea that, out of a point cloud map, LTS-Net can extract only those points which belong to temporal stable features such as poles and trunks, which are also the only features present in April's bag. Interestingly, also the localisation's performance while using May and June's bags has seen a consistent improvement: more specifically, the APE for June drastically dropped from $0.16$ to $0.07$. 

Besides the increased localisation performance, another important advantage brought by LTS-NET is the possibility to reduce the number of maps the robot needs to build before being deployed autonomously. This is a crucial aspect when targeting outdoor robotic applications, in particular navigating an orchard which can extend for multiple square kilometres, making mapping a tedious and expensive task. In our experiments, we found that out of the four maps used, having only two maps (e.g., March and June) would be sufficient for achieving a precise localisation across four months. This is already a positive result because guarantees there is no need for mapping the vineyard every time the robot needs to be used for data collection or inspection. In addition, we showed that LST-NET can come in handy in extracting only those features which are stable across time (such as poles and trunks), removing from the map those points belonging to grass and leaves. In this way, it is possible not only to improve the localisation accuracy in general but to further extend the time required between two mapping sessions, with a saving in time and cost. Unfortunately, we noted that it was not possible to achieve a successful localisation while using the ROS bag recorded in March while using a map built in June. Hence, future directions of this work will be towards improving the features extraction capabilities of LTS-NET by also integrating the additional data that is planned to be recorded until the end of the next harvesting season. 

\section{Conclusion}\label{sec:conclusion}
The present manuscript aims at discussing the problem of the long-term deployment of an autonomous robot in a continuously changing environment, such as a vineyard. To this scope, we established a robust and reproducible data recording pipeline, starting the data collection in March 2022 and with plans to continue until the end of the harvesting season in October 2022. Some of the data recording sessions have been used as a benchmark for testing mapping and localisation performance across time. The results show that re-using maps built previously in time produces a drop in performance in map-based localisation algorithms, mainly due to the natural seasonal changes of the plants and crops. However, we showed that by extracting stable features out of the maps, we can improve the localisation error and reduce the frequency at which news maps must be created for guaranteeing a correct and efficient deployment of the robotic platform.

\bibliographystyle{IEEEtran}
% \balance
\bibliography{IEEEexample}

\end{document}